\documentclass[conference]{IEEEtran}
\ifCLASSOPTIONcompsoc
\usepackage[nocompress]{cite}
\else
\usepackage{cite}
\fi

\ifCLASSINFOpdf
\else
\fi

\usepackage{amsmath,graphicx,float}
\usepackage{afterpage}
\usepackage{algorithmic,algorithm}
\usepackage{booktabs}
\usepackage{caption}

\graphicspath{ {images/} }
\begin{document}
%
\title{Deep Spiking Neural Network with \\ Spike Count based Learning Rule}
\IEEEoverridecommandlockouts 
\author{
	Jibin Wu$^1$, Yansong Chua$^2{^{*}}$\thanks{$^{*}$Corresponding Author: chuays@i2r.a-star.edu.sg}, Malu Zhang$^1$, Qu Yang$^1$, Guoqi Li$^3$, Haizhou Li$^1$ 
	\\
	\\$^1$Department of Electrical and Computer Engineering, National University of Singapore, Singapore
	\\$^2$Institute for Infocomm Research, A*STAR, Singapore
	\\$^3$Department of Precision Instrument, Beijing Innovation Center for Future Chip, Tsinghua University, China
}

\maketitle

\begin{abstract}
Deep spiking neural networks (SNNs) support asynchronous event-driven computation, massive parallelism and demonstrate great potential to improve the energy efficiency of its synchronous analog counterpart. However, insufficient attention has been paid to neural encoding when designing SNN learning rules. Remarkably, the temporal credit assignment has been performed on rate-coded spiking inputs, leading to poor learning efficiency. In this paper, we introduce a novel spike-based learning rule for rate-coded deep SNNs, whereby the spike count of each neuron is used as a surrogate for gradient backpropagation. We evaluate the proposed learning rule by training deep spiking multi-layer perceptron (MLP) and spiking convolutional neural network (CNN) on the UCI machine learning and MNIST handwritten digit datasets. We show that the proposed learning rule achieves state-of-the-art accuracies on all benchmark datasets. The proposed learning rule allows introducing latency, spike rate and hardware constraints into the SNN learning, which is superior to the indirect approach in which conventional artificial neural networks are first trained and then converted to SNNs.  Hence, it allows direct deployment to the neuromorphic hardware and supports efficient inference. Notably, a test accuracy of 98.40\% was achieved on the MNIST dataset in our experiments with only 10 simulation time steps, when the same latency constraint is imposed during training. 
\end{abstract}
\section{INTRODUCTION}
Deep learning has made remarkable progress in recent years, with huge impacts on many aspects of our daily lives \cite{hu2018spiking}. While being brain-inspired, deep learning models differ significantly from the biological brain in many ways. In human brain, the information is represented and communicated through asynchronous action potentials or spikes. To faithfully describe the dynamics of biological neural networks, several spiking neuron models have been proposed with different degree of biological realism\cite{gerstner2002spiking}. Although, how information is encoded and exchanged within networks of spiking neurons remain largely unknown, the inherent properties of spiking neural networks (e.g., low power event-driven computation and massive parallelism) have motivated a growing body of research works in the energy efficient neuromorphic hardware as well as compatible spike-based learning rules\cite{pfeiffer2018deep,som-snn}. 

Early studies of SNNs were focused mostly on a single layer of neurons, which establish a strong theoretical foundation in the neural coding and synaptic plasticity \cite{maass1997networks,gerstner2002spiking,dayan2001theoretical}. Motivated by the recent success in deep learning, research attention in SNNs has been shifted towards networks with multiple hidden layers \cite{tavanaei2018deep,pfeiffer2018deep}. However, training deep SNNs remains a challenging task due to the non-differentiability of spike generation. To overcome this, differentiable proxies have been employed to enable the powerful gradient backpropagation algorithm, examples include the membrane potential\cite{lee2016training,shrestha2018slayer,wu2018direct}, spike timing of first spike \cite{mostafa2018supervised} and spike statistics \cite{o2016deep,stromatias2017event}.  

While much progress has been made on spike-based learning rules in recent years, we observe that comparatively less attention has been paid to how information is represented in the network (i.e., neural encoding)  while developing these learning rules. Specifically, we argue that the temporal credit assignment is unnecessary when sensory inputs are rate encoded\cite{ratecode}, whereby spike timing carries no additional information. The problem is amplified when using traditional computer vision datasets or their neuromorphic versions to benchmark novel spike-based learning rules, wherein negligible time information exists \cite{iyer2018neuromorphic}. 

Another line of research in deep SNN involves the conversion of pre-trained ANNs to SNNs of the same network architecture\cite{cao2015spiking,diehl2015fast,ethImageNet,sengupta2018going,hu2018spiking}. This indirect training approach assumes the graded activation of analog neurons is equivalent to the average firing rate of spiking neurons, and simply requires parsing and normalizing the weights after training the ANNs. Notably, Rueckauer et al. provide a theoretical analysis of the performance deviation of such approach as well as a systematic study of frequently used layers in the CNN \cite{ethImageNet,sengupta2018going,hu2018spiking}. This conversion approach achieves the best-reported results for SNNs on many benchmark datasets including the challenging ImageNet dataset\cite{deng2009imagenet}. Nevertheless, the latency and accuracy trade-off has been identified as the main shortcoming of such an approach\cite{diehl2015fast}, requiring additional techniques to improve the latency and power efficiency \cite{neil2016learning}.

In this paper, to effectively process the rate-coded sensory inputs and feature vectors with a deep SNN, we propose a novel spike-based learning rule based on the non-leaky integrate-and-fire (IF) neuron. The temporal information associated with spikes is ignored in such a neuron model. Moreover, the non-differentiability of spike generation is circumvented by the use of spike count as a surrogate for gradient backpropagation. In contrast to the indirect conversion approach, the proposed rule uses the spike count information that can be directly obtained from spiking neurons. In addition, the latency, spike rate and other hardware constraints can be incorporated during the training phase, allowing direct deployment and efficient inference on the neuromorphic hardware.

The rest of this paper is organized as follows: in Section II, we present the proposed spike-based learning rule. In Section III, we evaluate the proposed learning rule on the UCI machine learning and MNIST benchmark datasets. Finally, we conclude with a further discussion in Section IV.

\vspace{5mm}
\section{Methods}
\label{ssec:Methods}
\subsection{Neuron Model}
In this work, we use the integrate-and-fire (IF) neuron model. This model faithfully retains the number of input spikes it receives (until reset) and its output spike count is independent of the spike timing of its inputs. While the IF neuron does not emulate the rich temporal dynamics of biological neurons, it is however ideal for working with rate-coded sensory input where spike timings don't play a role. 

At each time step $t$, the input spikes to neuron $j$ at layer $l$ are integrated as follows
\begin{equation}
z_j^l(t) = \vartheta  \cdot \sum\nolimits_i {w_{ji}^{l - 1} \cdot \theta _i^{l - 1}} (t)\\
\end{equation}
where $\vartheta$ is the neuron firing threshold and $\theta _i^{l - 1}(t)$ indicates the occurrence of an input spike from afferent neuron $i$ at time step $t$. The $w_{ji}^{l - 1}$ denotes the synaptic weight that connects afferent neuron $i$ from layer $l-1$.

Neuron $j$ then integrates the input current $z_j^l(t)$ into its membrane potential $V_j^l(t)$ as per Eq. 2. $V_j^l(t)$ is initialized with a learnable parameter ${b_j}$ (Eq. 3), and an output spike is generated whenever $V_j^l(t)$ crosses the firing threshold $\vartheta$ (Eq. 4).

\begin{equation}
V_j^l(t) = V_j^l(t - 1) + z_j^l(t) - \vartheta \cdot \theta _j^l(t - 1)
\end{equation}

\begin{equation}
V_j^l(0) =  {b_j}
\end{equation}

\begin{equation}
\theta _j^l(t) = \Theta (V_j^l(t) - \vartheta) \;\; with \;\; \Theta (x) = \left\{\begin{array}{l}1,\;\;\; if\;x \ge 0\\
0,\;\;\;otherwise\;
\end{array} \right.
\end{equation}

The total number of spikes (i.e., spike count) generated by neuron $i$ at the input layer can be determined by summing all incoming spikes over the simulation period $T$ as per Eq. 5. For static image inputs, both raw intensity values or aggregate spike counts from a Poisson generator can be used as the input.
\begin{equation}
a_i^0 = \sum\nolimits_t^T {\theta _i^0(t)} \\
\end{equation}

\begin{figure}[htb]	
	\begin{minipage}[b]{1.0\linewidth}
		\centering
		\centerline
		{\includegraphics[width = 8 cm]{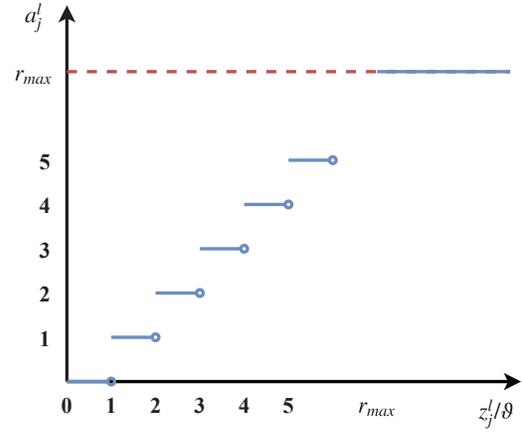}}
		\caption{Illustration of the transfer function for the spike count $a_j^l$. The $a_j^l$ is determined by rounding  ${z_j^l}\slash{\vartheta}$  towards nearest integer and also upper bounded by the maximum firing rate $r_{max}$.}
		\label{fig:activationFunc}
	\end{minipage}
\end{figure}

According to Eq. 1, the aggregated input current of neuron $j$ in layer $l$ can be expressed as
\begin{equation}
z_j^l = \vartheta  \cdot \sum\nolimits_i {w_{ji}^{l - 1} \cdot a_i^{l - 1}} + b_j^l \\
\end{equation}
where $a_i^{l - 1}$ is the input spike count from pre-synaptic neuron $i$ at layer $l-1$ and $b_j^l$ is the initial membrane potential of post-synaptic neuron $j$ at layer $l$.

Different from the continuous neuron activation function that used in the traditional ANNs,
output spike counts are only non-negative integers (enforced by the term  $z_j^l > 0$ in Eq. 7). The surplus membrane potential that insufficient to induce an additional spike is ignored for the next sample as shown in Fig. \ref{fig:activationFunc} and Eq. 7. Such rounding effort leads to a quantization error, which can be compensated by normalizing the synaptic weights with zero mean in the subsequent layer. Moreover, the output spike counts are upper bounded by the maximum time steps $r_{max} := T/dt$, such constraint can be alleviated using a higher time resolution $dt$. In practice, we have not noticed any performance drop due to $r_{max}$ from our experimental results on the UCI and MNIST datasets.

\begin{equation}
\begin{aligned}
a_j^l &= f(z_j^l) \\
&= clamp\left(\left\lfloor {\frac{{z_j^l}}{{\vartheta}}} \right\rfloor \cdot\left( {z_j^l > 0} \right), r_{max} \right)\\
&= min\left(\left\lfloor {\frac{{z_j^l}}{{\vartheta}}} \right\rfloor \cdot\left( {z_j^l > 0} \right), r_{max} \right)
\end{aligned}
\end{equation}
where the output spike count will be clipped at a value of zero for negative aggregated input current $z_j^l$.

\subsection{Back-propagation in Rate-coded Deep SNNs}
Here, we derive the backpropagation algorithm using the spike count as a surrogate for gradient propagation.

\subsubsection{Loss Function}
In this work, the Cross-Entropy loss function that is commonly used for classification tasks is employed as per Eq. 8, which transforms the real-valued outputs to a normalized probability distribution. Other loss functions used in the ANNs may also be applied.
\begin{equation}
\begin{aligned}
E(a_j^{{n_l}},{y_j}) &=  - \log \left( {\frac{{\exp \left( {a_j^{{n_l}}} \right)}}{{\sum\nolimits_k {\exp \left( {a_k^{{n_l}}} \right)} }}}\right)\\
\\&= \log \left( {\sum\nolimits_k {\exp (a_k^{{n_l}})} } \right) - a_j^{{n_l}}
\end{aligned}
\end{equation}
where $k$ refers to neurons at the output layer.

The partial derivative of Cross-Entropy loss with respect to the output spike count can be determined as
\begin{equation}
\frac{{\partial E}}{{\partial a_j^{{n_l}}}} = \frac{{\exp \left( {a_j^{{n_l}}} \right)}}{{\sum\nolimits_k {\exp \left( {a_k^{{n_l}}} \right)} }} - {y_j}
\end{equation}
where $a_j^{n_l}$ is the output spike count and $y_j$ is the desired one-hot label for neuron $j$ at output layer $n_l$.

\subsubsection{Output Layer}
Following Eqs. 6, 7 and 9, the partial derivatives of the loss function with respect to the synaptic weight $w_{ji}^{{n_l} - 1}$ and bias term $b_j^{{n_l}}$ can be expressed in Eqs. 10 and 11, respectively. As per common practice, we denote the term ${\partial E/\partial z_j^{{n_l}}} = \delta _j^{{n_l}}$.
\begin{equation}
\begin{aligned}
\frac{{\partial E}}{{\partial w_{ji}^{{n_l} - 1}}} &= \frac{{\partial E}}{{\partial z_j^{{n_l}}}}\frac{{\partial z_j^{{n_l}}}}{{\partial w_{ji}^{{n_l} - 1}}}\\
\\&= \frac{{\partial E}}{{\partial a_j^{{n_l}}}}\frac{{\partial a_j^{{n_l}}}}{{\partial z_j^{{n_l}}}}\frac{{\partial z_j^{{n_l}}}}{{\partial w_{ji}^{{n_l} - 1}}}\\
\\&= \underbrace {\left( {\frac{{\exp \left( {a_j^{{n_l}}} \right)}}{{\sum\nolimits_k {\exp \left( {a_k^{{n_l}}} \right)} }} - {y_j}} \right)\left( {\frac{1}{\vartheta } \cdot \left( {z_j^{{n_l}} > 0} \right)} \right)}_{\delta _j^{{n_l}}}
\\&\ \ \ \ \ \cdot \left( {\vartheta  \cdot a_i^{{n_l} - 1}} \right)
\end{aligned}
\end{equation}

\begin{equation}
\frac{{\partial E}}{{\partial b_j^{{n_l}}}} = \frac{{\partial E}}{{\partial z_j^{{n_l}}}}\frac{{\partial z_j^{{n_l}}}}{{\partial b_j^{{n_l}}}} = \delta _j^{{n_l}}
\end{equation}

\subsubsection{Hidden Layers}
Similar to Eqs. 10 and 11, the partial derivatives of the loss function with respect to the synaptic weight $w_{ji}^{l-1}$ and bias term ${b_j^l}$ for hidden layer $l$ can be expressed in Eqs. 12 and 13 below.

\begin{equation}
\begin{aligned}
\frac{{\partial E}}{{\partial w_{ji}^{l-1}}} &= \frac{{\partial E}}{{\partial z_j^{l}}}\frac{{\partial z_j^{l}}}{{\partial w_{ji}^{l-1}}}\\
&= \delta _j^{l} \cdot \left({\vartheta  \cdot a_i^{l-1}} \right)
\end{aligned}
\end{equation}

\begin{equation}
\frac{{\partial E}}{{\partial b_j^l}} = \frac{{\partial E}}{{\partial z_j^l}}\frac{{\partial z_j^l}}{{\partial b_j^l}} = \delta _j^l
\end{equation}
where
\begin{equation}
\begin{aligned}
\delta _j^l &= \frac{{\partial E}}{{\partial z_j^l}}\\
& = \sum\nolimits_k {\frac{{\partial E}}{{\partial z_k^{l + 1}}}} \frac{{\partial z_k^{l + 1}}}{{\partial z_j^l}}\\
& = \sum\nolimits_k {\delta _k^{l + 1}} \frac{{\partial z_k^{l + 1}}}{{\partial a_j^l}}\frac{{\partial a_j^l}}{{\partial z_j^l}}\\
& = \sum\nolimits_k {\delta _k^{l + 1}} \left( {\vartheta  \cdot w_{kj}^l} \right)\left( {\frac{1}{{\vartheta}} \cdot \left( {z_j^l > 0} \right)} \right)\\
\end{aligned}
\end{equation}

Such a direct training approach allows easy integration of hardware constraints into the loss function and optimized jointly during training, including spike rate, inference latency and limited synaptic weight precision etc. Hence, facilitating more convenient deployment and better inference performance on the real neuromorphic hardware.

\begin{table*}[!htb]{\tiny}
	\centering
	\caption{Details of experimental setup and classification accuracy of selected benchmark datasets from UCI machine learning repository. The results are averaged over 5 experimental runs with random weight initialization.} 
	\resizebox{18 cm}{!}{
		\begin{tabular}{lccccccc}						
			\toprule
			\bottomrule 
			\textbf{Dataset} &\textbf{Tr}  & \textbf{Ts}  & \textbf{Features} & \textbf{Classes} & \textbf{Network Structure} & \begin{tabular}{c}\textbf{Accuracy (Tr/Te)} \\\textbf{in this work (\%)}\end{tabular} & \begin{tabular}{c}\textbf{Accuracy (Tr/Te)} \\\textbf{in} \cite{wang2017spiketemp} \textbf{(\%)} \end{tabular}\\ 
			\hline  
			\rule{0pt}{2ex}Iris  & 90  & 60  & 4  & 3 & 4-20-3 & 100/100 & 100/96.7\\
			WBC  & 455  & 228  & 9  & 2 & 9-20-2 & 100/100 & 99.1/98.3\\
			Abalone  & 2000  & 2177  & 7  & 3 & 8-50-2 & 100/100 & 45.7/47.8\\
			Yeast  & 990  & 494  & 8  & 10 & 8-50-10 & 100/100 & 56.7/31.6\\
			\toprule
			\bottomrule
		\end{tabular}
		\label{table:result}
	}
\end{table*}

\vspace{5mm}
\section{EXPERIMENTAL RESULTS}
In this section, we evaluate the proposed spike count based learning rule on the traditional machine learning and image classification tasks. 
\subsection{UCI Classification Tasks}
To evaluate rate-coded SNN models, we use datasets from the UCI machine learning repository that have been widely used for benchmarking machine learning and neural network models\cite{uci}. The following four datasets are used: 1) Iris; 2) Wisconsin Breast Cancer (WBC); 3) Abalone; 4) Yeast. For a fair comparison, the experimental setups follow those from recent work on the rank-order learning for SNN\cite{wang2017spiketemp}.  Table I summarizes the experimental setups and classification results for each dataset: 1) the splitting of training (Tr) and testing (Te) samples; 2) the number of features; 3) the number of output classes; 4) the network structure used for each dataset, and 5) classification accuracies for train and test set.

The input feature vectors are normalized within [0,1], thereafter Poisson spike trains are generated for each feature dimension with firing rates proportional to the normalized feature value. The simulation period of $T$ = 20 ms with a simulation time step of 1 ms (i.e., $r_{max} = 20$  Hz) is used. We initialize the SNN by setting firing threshold $\vartheta$ and learning rate $\lambda$ to 1.0 and $5*10^{-4}$, respectively. The weights for the SNN classifier are drawn randomly from a Gaussian distribution with a mean of 0 and standard deviation of 0.05. Adam optimizer\cite{kingma2014adam} is used for parameter update. For each network structure, 5 SNNs with random weight initialization are trained and the average classification results are reported.

As shown in Table I, the deep SNN trained with the proposed learning rule achieves 100\% accuracies consistently for all four benchmark datasets. In contrast, the SNN trained with rank-order learning \cite{wang2017spiketemp} achieves only competitive results for the easier Iris and WBC datasets, while the test accuracies degrade significantly to less than 50\% for the more challenging Abalone and Yeast datasets. Although rank-order learning generally implies low latency and low spike rates, it is worth noting that it only applies to single-layer networks, whereby the input encoding layer is directly connected to the output layer. Therefore, the representation powers of these SNN models are greatly limited. In contrast, the proposed learning rule overcomes this limitation and can scale well with multiple hidden layers.

\subsection{MNIST Classification Task}
We further evaluate our proposed learning rule using the standard MNIST dataset of handwritten digits that is widely used for benchmarking multi-layer SNN learning rules \cite{tavanaei2018deep}. The training and testing sets consist of 60,000 and 10,000 grayscale images of 28 $\times$ 28 pixels. Similar to the experimental setup used for UCI datasets, the input spike trains are generated from a Poisson generator, whereby firing rates are proportional to the normalized pixel intensity. The simulation period of $T$ = 50 ms with a simulation time step of 1 ms (i.e., $r_{max} = 50$ ) is used. We initialize the SNN by setting the firing threshold $\vartheta$ and learning rate $\lambda$ to 1.0 and $10^{-3}$, respectively. We perform all the experiments using the Pytorch library, whereby the dynamics of the IF neuron as mentioned in Section. \ref{ssec:Methods} are explicitly modeled during training and testing. The weights are initialized with default values in Pytorch, and we use the Adam optimizer for parameter update. For each network structure, 5 SNNs with random weight initialization are trained and the average classification results are reported.

We perform experiments using two common feedforward neural network architectures: the multi-layer perceptron (MLP) and convolutional neural network (CNN). For the MLP, we explore the use of two network structures (describe in terms of the number of neurons in each layer): 784-800-10 and 784-800-800-10. As shown in Table II, the SNN models trained with the proposed learning rule achieves classification accuracies of 98.64\% and 98.66\% for one and two hidden layers, respectively. These accuracies are competitive with both spike-based learning rules \cite{o2016deep,lee2016training,neftci2017event,mostafa2018supervised,spatiotemporal} and ANN conversion approaches\cite{diehl2015fast,neil2016learning} as summarized in Table II. 

CNNs are currently the default choice for many computer vision tasks, including image classification \cite{krizhevsky2012imagenet}, detection\cite{vggnet} and segmentation\cite{redmon2016you}. For SNNs, the best reported result for the MNIST dataset also employs a CNN architecture\cite{ethImageNet}. Here, we apply the proposed learning rule to train a spiking-CNN with the CNN architecture of 28$\times$28-12c5-2a-64c5-2a-10. The notation `12c5' denotes 12 convolution kernel of size 5 $\times$ 5 and `2a' denotes average pooling of size 2 $\times$ 2. The outputs from the final average pooling layer are vectorized and fully connected to the output layer. As shown in Table II, the spiking-CNN model trained with the proposed rule offers a promising classification accuracy of 99.26\%. It also worth mentioning that neither additional data augmentation nor advanced techniques such as batch normalization or dropout are applied in this work; we expect the accuracies to be further improved when these techniques are applied. 

We note that many existing spike-based learning rules for deep SNN consider the spike timing as useful information. Despite promising results achieved with these rules on standard benchmark datasets such as MNIST and CIFAR-10, we expect longer training time and more memory to compute and store the dynamics of neuron than the proposed learning rule.

\begin{table*}[!htp]
	\centering	
	\huge
	\caption{Comparison of classification accuracies of deep SNNs trained with the proposed and other supervised learning rules on the MNIST dataset (For more details, refer to the review paper\cite{tavanaei2018deep}).} 				
	\resizebox{18 cm}{!}{
		\begin{tabular}{lccc}
			\toprule
			\bottomrule 
			\textbf{Model} & \textbf{Network Architecture} &\textbf{Method} & \textbf{Test Accuracy (\%)} \\ 
			\toprule
			O'Connor (2016) \cite{o2016deep} & MLP &Fractional stochastic gradient descent & 97.93\\
			Lee (2017) \cite{lee2016training}  & MLP & Backpropagation & 98.88 \\
			Neftci (2017) \cite{neftci2017event}  & MLP & Event-driven random backpropagation & 97.98 \\
			Mostafa (2017) \cite{mostafa2018supervised} & MLP &  Backpropagation with temporal coding & 98.00 \\
			Wu (2018)   \cite{spatiotemporal}  & MLP & Spatio-Temporal Backpropagation  & 98.48 \\
			Diehl (2015) \cite{diehl2015fast}  & MLP & Conversion of ANNs  & 98.60 \\
			Neil (2016) \cite{neil2016learning}  & MLP & Conversion of ANNs  & 98.00 \\
			\textbf{This work} & MLP (784-800-10) & Backpropagation with rate-coded SNN &\textbf{98.64} \\
			\textbf{This work} & MLP (784-800-800-10) & Backpropagation with rate-coded SNN & \textbf{98.66} \\
			\toprule
			Lee (2017) \cite{lee2016training}  & CNN & Backpropagation & 99.31 \\
			Shrestha (2018) \cite{shrestha2018slayer}  & CNN & Backpropagation  & 99.36 \\
			Diehl (2015) \cite{diehl2015fast}  & CNN & Conversion of ANNs  & 99.10 \\
			Rueckauer (2017) \cite{ethImageNet}  & CNN & Conversion of ANNs  & 99.44 \\
			Kheradpisheh (2018) \cite{kheradpisheh2018stdp}  & CNN & Layerwise STDP + SVM  & 98.40 \\
			\textbf{This work} & CNN & Backpropagation with rate-coded SNN &\textbf{99.26} \\
			\toprule
			\bottomrule
		\end{tabular}
	}
\end{table*}

The latency and accuracy trade-off have been identified for the indirect ANN conversion approach, whereby classification accuracy improves over time when more evidence is accumulated \cite{diehl2015fast}. Although techniques \cite{neil2016learning} have been proposed to effectively improve the latency and power efficiency, they generally require more training time and hyperparameter tuning. In our approach, however, the latency and other hardware constraints are integrated during the training phase of the proposed learning rule, allowing direct deployment to the neuromorphic hardware for efficient inference without additional work as proposed for the indirect conversion approach\cite{neil2016learning}. For instance, to reduce the inference time, we can explicitly constraint the simulation period with $T$ = 10 ms for both training and testing. Notably, the MLP model (784-800-10) is able to achieve a classification accuracy of 98.40\%, which is quite close to the accuracy when trained with $T$ = 50 ms for the MNIST dataset.

\section{Discussion and Conclusion}
Motivated by the fact that no useful temporal information is encoded in spike timing for rate-code spiking inputs, we introduce a novel spike-based learning rule to train deep SNNs, whereby the spike count of each neuron is used as the surrogate for gradient backpropagation. Differing from other spike-based learning rules, which consider the spike timing during error backpropagation\cite{lee2016training, shrestha2018slayer}, the proposed learning rule requires much lesser computation and memory. Moreover, the proposed learning rule demonstrates competitive classification accuracies on both UCI machine learning and MNIST datasets. 

In contrast to the indirect ANN to SNN conversion approach, the proposed learning rule can integrate the inference latency, spike rate and hardware constraints more effectively during the training. Hence, it allows direct deployment to neuromorphic hardware for efficient inference. Despite promising results are achieved on the MNIST dataset, the quantization error as shown in the surplus membrane potential of spiking neurons may become severe when these errors are accumulated over many layers. In future work, we will investigate how to scale up the learning rule to deeper neural network architectures, such as VGGNet and ResNet, so as to solve more challenging tasks. 

\section*{Acknowledgments}
This research is supported by Programmatic grant no. A1687b0033 from the Singapore Government's Research, Innovation and Enterprise 2020 plan (Advanced Manufacturing and Engineering domain)

\bibliographystyle{IEEEbib}
\bibliography{citation-main}

\begin{thebibliography}{10}

\bibitem{lecun2015deep}
Y.~LeCun, Y.~Bengio, and G.~Hinton,
\newblock ``Deep learning,''
\newblock {\em Nature}, vol. 521, no. 7553, pp. 436, 2015.

\bibitem{gerstner2002spiking}
W.~Gerstner and W.~M. Kistler,
\newblock {\em Spiking neuron models: Single neurons, populations, plasticity},
\newblock Cambridge University Press, 2002.

\bibitem{pfeiffer2018deep}
M.~Pfeiffer and T.~Pfeil,
\newblock ``Deep learning with spiking neurons: Opportunities \& challenges,''
\newblock {\em Frontiers in Neuroscience}, vol. 12, pp. 774, 2018.

\bibitem{som-snn}
J.~Wu, Y.~Chua, M.~Zhang, H.~Li, and K.~C. Tan,
\newblock ``A spiking neural network framework for robust sound
  classification,''
\newblock {\em Frontiers in Neuroscience}, vol. 12, pp. 836, 2018.

\bibitem{maass1997networks}
W.~Maass,
\newblock ``Networks of spiking neurons: the third generation of neural network
  models,''
\newblock {\em Neural networks}, vol. 10, no. 9, pp. 1659--1671, 1997.

\bibitem{dayan2001theoretical}
P.~Dayan and L.~F. Abbott,
\newblock {\em Theoretical neuroscience}, vol. 806,
\newblock Cambridge, MA: MIT Press, 2001.

\bibitem{tavanaei2018deep}
A.~Tavanaei, M.~Ghodrati, S.~R. Kheradpisheh, T.~Masquelier, and A.~S. Maida,
\newblock ``Deep learning in spiking neural networks,''
\newblock {\em arXiv preprint arXiv:1804.08150}, 2018.

\bibitem{lee2016training}
J.~H. Lee, T.~Delbruck, and M.~Pfeiffer,
\newblock ``Training deep spiking neural networks using backpropagation,''
\newblock {\em Frontiers in Neuroscience}, vol. 10, pp. 508, 2016.

\bibitem{shrestha2018slayer}
S.~B. Shrestha and G.~Orchard,
\newblock ``Slayer: Spike layer error reassignment in time,''
\newblock {\em arXiv preprint arXiv:1810.08646}, 2018.

\bibitem{wu2018direct}
Y.~Wu, L.~Deng, G.~Li, J.~Zhu, and L.~Shi,
\newblock ``Direct training for spiking neural networks: Faster, larger,
  better,''
\newblock {\em arXiv preprint arXiv:1809.05793}, 2018.

\bibitem{mostafa2018supervised}
H.~Mostafa,
\newblock ``Supervised learning based on temporal coding in spiking neural
  networks,''
\newblock {\em IEEE Transactions on Neural Networks and Learning Systems}, vol.
  29, no. 7, pp. 3227--3235, 2018.

\bibitem{o2016deep}
P.~O'Connor and M.~Welling,
\newblock ``Deep spiking networks,''
\newblock {\em arXiv preprint arXiv:1602.08323}, 2016.

\bibitem{stromatias2017event}
E.~Stromatias, M.~Soto, T.~Serrano-Gotarredona, and B.~Linares-Barranco,
\newblock ``An event-driven classifier for spiking neural networks fed with
  synthetic or dynamic vision sensor data,''
\newblock {\em Frontiers in neuroscience}, vol. 11, pp. 350, 2017.

\bibitem{ratecode}
E.~N. Brown, R.~E. Kass, and P.~P. Mitra,
\newblock ``Multiple neural spike train data analysis: state-of-the-art and
  future challenges,''
\newblock {\em Nature Neuroscience}, vol. 7, no. 5, pp. 456, 2004.

\bibitem{iyer2018neuromorphic}
L.~R. Iyer, Y.~Chua, and H.~Li,
\newblock ``Is neuromorphic mnist neuromorphic? analyzing the discriminative
  power of neuromorphic datasets in the time domain,''
\newblock {\em arXiv preprint arXiv:1807.01013}, 2018.

\bibitem{cao2015spiking}
Y.~Cao, Y.~Chen, and D.~Khosla,
\newblock ``Spiking deep convolutional neural networks for energy-efficient
  object recognition,''
\newblock {\em International Journal of Computer Vision}, vol. 113, no. 1, pp.
  54--66, 2015.

\bibitem{diehl2015fast}
P.~U. Diehl, D.~Neil, J.~Binas, M.~Cook, S.~C. Liu, and M.~Pfeiffer,
\newblock ``Fast-classifying, high-accuracy spiking deep networks through
  weight and threshold balancing,''
\newblock in {\em 2015 International Joint Conference on Neural Networks
  (IJCNN)}, July 2015, pp. 1--8.

\bibitem{ethImageNet}
B.~Rueckauer, I.~A. Lungu, Y.~Hu, M.~Pfeiffer, and S.~C. Liu,
\newblock ``Conversion of continuous-valued deep networks to efficient
  event-driven networks for image classification,''
\newblock {\em Frontiers in Neuroscience}, vol. 11, pp. 682, 2017.

\bibitem{sengupta2018going}
A.~Sengupta, Y.g Ye, C.~Wang, R.and~Liu, and K.~Roy,
\newblock ``Going deeper in spiking neural networks: Vgg and residual
  architectures,''
\newblock {\em arXiv preprint arXiv:1802.02627}, 2018.

\bibitem{hu2018spiking}
Y.~Hu, H.~Tang, Y.~Wang, and G.~Pan,
\newblock ``Spiking deep residual network,''
\newblock {\em arXiv preprint arXiv:1805.01352}, 2018.

\bibitem{deng2009imagenet}
J.~Deng, W.~Dong, R.~Socher, L.~Li, Kai Li, and Li~Fei-Fei,
\newblock ``Imagenet: A large-scale hierarchical image database,''
\newblock in {\em 2009 IEEE Conference on Computer Vision and Pattern
  Recognition}, June 2009, pp. 248--255.

\bibitem{neil2016learning}
D.~Neil, M.~Pfeiffer, and S.~C. Liu,
\newblock ``Learning to be efficient: algorithms for training low-latency,
  low-compute deep spiking neural networks,''
\newblock in {\em Proceedings of the 31st Annual ACM Symposium on Applied
  Computing}. ACM, 2016, pp. 293--298.

\bibitem{wang2017spiketemp}
J.~Wang, A.~Belatreche, L.~P. Maguire, and T.~M. McGinnity,
\newblock ``Spiketemp: An enhanced rank-order-based learning approach for
  spiking neural networks with adaptive structure,''
\newblock {\em IEEE Transactions on Neural Networks and Learning Systems}, vol.
  28, no. 1, pp. 30--43, 2017.

\bibitem{uci}
A.~Asuncion and D.~Newman,
\newblock ``Uci machine learning repository,'' 2007.

\bibitem{kingma2014adam}
D.~P. Kingma and J.~Ba,
\newblock ``Adam: A method for stochastic optimization,''
\newblock {\em arXiv preprint arXiv:1412.6980}, 2014.

\bibitem{neftci2017event}
E.~O. Neftci, C.~Augustine, S.~Paul, and G.~Detorakis,
\newblock ``Event-driven random back-propagation: Enabling neuromorphic deep
  learning machines,''
\newblock {\em Frontiers in neuroscience}, vol. 11, pp. 324, 2017.

\bibitem{spatiotemporal}
Y.~Wu, L.~Deng, G.~Li, J.~Zhu, and L.~Shi,
\newblock ``Spatio-temporal backpropagation for training high-performance
  spiking neural networks,''
\newblock {\em Frontiers in Neuroscience}, vol. 12, pp. 331, 2018.

\bibitem{krizhevsky2012imagenet}
A.~Krizhevsky, I.~Sutskever, and G.~E. Hinton,
\newblock ``Imagenet classification with deep convolutional neural networks,''
\newblock in {\em Advances in Neural Information Processing Systems}, 2012, pp.
  1097--1105.

\bibitem{vggnet}
K.~Simonyan and A.~Zisserman,
\newblock ``Very deep convolutional networks for large-scale image
  recognition,''
\newblock {\em arXiv preprint arXiv:1409.1556}, 2014.

\bibitem{redmon2016you}
J.~Redmon, S.~Divvala, R.~Girshick, and A.~Farhadi,
\newblock ``You only look once: Unified, real-time object detection,''
\newblock in {\em Proceedings of the IEEE conference on computer vision and
  pattern recognition}, 2016, pp. 779--788.

\bibitem{kheradpisheh2018stdp}
S.~R. Kheradpisheh, M.~Ganjtabesh, S.~J. Thorpe, and T.~Masquelier,
\newblock ``Stdp-based spiking deep convolutional neural networks for object
  recognition,''
\newblock {\em Neural Networks}, vol. 99, pp. 56--67, 2018.

\end{thebibliography}
\end{document}